\documentclass[conference,a4paper]{IEEEtran}
\usepackage{xspace}
\newcommand{\eg}{e.g.\xspace}

\usepackage{cite}
\usepackage{amsmath,amssymb,amsfonts}
\usepackage{graphicx}
\usepackage{booktabs}
\usepackage{siunitx}
\usepackage{multirow}
\usepackage{array}
\usepackage{balance}
\usepackage{algorithm}
\usepackage[hidelinks]{hyperref}

\usepackage{hyperref}
\usepackage{algorithmic}
\usepackage{textcomp}
\def\BibTeX{{\rm B\kern-.05em{\sc i\kern-.025em b}\kern-.08em
    T\kern-.1667em\lower.7ex\hbox{E}\kern-.125emX}}

\usepackage{fancyhdr}
\fancypagestyle{firstpage}{%
    \fancyhf{} 
    \fancyhead[L]{2025 28th International Conference on Computer and Information Technology (ICCIT) \\ 19-21 December 2025, Cox's Bazar, Bangladesh}
    \fancyfoot[L]{\copyright 2025 IEEE. Personal use of this material is permitted. Please cite the published version from IEEE Xplore. \hfill}
}

\begin{document}
\title{\huge Fixed-Threshold Evaluation of a Hybrid CNN–ViT for AI-Generated Image Detection Across Photos and Art}


\author{
    \IEEEauthorblockN{
        Md Ashik Khan\textsuperscript{1} and 
        Arafat Alam Jion\textsuperscript{2}
    }
    \IEEEauthorblockA{
        \textit{\textsuperscript{1}Department of Computer Science and Engineering, Indian Institute of Technology Kharagpur, India}\\
        \textit{\textsuperscript{2}Department of Electrical and Electronic Engineering, Chittagong University of Engineering and Technology, Bangladesh}
    }
    \IEEEauthorblockA{
        aasshhik98@gmail.com, arafatjion2025@gmail.com
    }
}

\maketitle
\thispagestyle{firstpage}

\begin{abstract}
AI image generators create both photorealistic images and stylized art, necessitating robust detectors that maintain performance under common post-processing transformations (JPEG compression, blur, downscaling). Existing methods optimize single metrics without addressing deployment-critical factors such as operating point selection and fixed-threshold robustness. This work addresses misleading robustness estimates by introducing a fixed-threshold evaluation protocol that holds decision thresholds, selected once on clean validation data, fixed across all post-processing transformations. Traditional methods retune thresholds per condition, artificially inflating robustness estimates and masking deployment failures. We report deployment-relevant performance at three operating points (Low-FPR, ROC-optimal, Best-F1) under systematic degradation testing using a lightweight CNN-ViT hybrid with gated fusion and optional frequency enhancement. Our evaluation exposes a statistically validated forensic-semantic spectrum: frequency-aided CNNs excel on pristine photos but collapse under compression (93.33\% to 61.49\% accuracy at JPEG Q60), whereas ViTs degrade minimally (92.86\% to 88.36\%) through robust semantic pattern recognition. Multi-seed experiments (N=3, 95\% CI) demonstrate that all architectures achieve 15\% higher AUROC on artistic content (0.901--0.907) versus photorealistic images (0.747--0.759), confirming that semantic patterns provide fundamentally more reliable detection cues than forensic artifacts. Our hybrid approach achieves balanced cross-domain performance: 91.4\% accuracy on tiny-genimage photos, 89.7\% on AiArtData art/graphics, and 98.3\% (competitive) on CIFAKE. Fixed-threshold evaluation eliminates retuning inflation, reveals genuine robustness gaps, and yields actionable deployment guidance: prefer CNNs for clean photo verification, ViTs for compressed social media content, and hybrids for art/graphics screening.
\end{abstract}


\section{Introduction}
\label{sec:intro}
Text-to-image diffusion models~\cite{ho2020ddpm,rombach2022ldm} and adversarial generators~\cite{karras2019stylegan,karras2021stylegan3} enable widespread creation of photorealistic photographs and stylized artwork, simultaneously unlocking creativity while eroding
trust in visual media. Newsrooms, e-commerce platforms, and social networks encounter increasing volumes of AI-generated content ranging from benign artistic expression to malicious disinformation. While cryptographic watermarking and provenance solutions provide complementary authentication signals, they are frequently stripped during cross-platform distribution or absent from legacy content, necessitating robust content-based detection systems.

A critical gap exists between academic evaluation protocols and deployment requirements. Current approaches predominantly optimize averaged metrics (\eg, AUROC) without considering application-specific operating constraints: newsroom verification systems require ultra-low false-positive rates, while content moderation platforms prioritize balanced accuracy. More fundamentally, conventional robustness assessments employ \emph{retuned thresholds}: decision boundaries are re-optimized per distortion condition, artificially inflating reported performance and obscuring failures that occur when a single threshold is deployed system-wide. Cross-domain evaluation remains underexplored, particularly the distinction between photorealistic images that stress low-level forensic artifacts and artistic content that emphasizes global semantic patterns.

We address these evaluation gaps through a systematic study employing fixed-threshold robustness protocols across deployment-relevant operating points. Using a gated CNN-ViT fusion architecture with optional frequency enhancement, we evaluate performance at three operating points (Low-FPR (\(\sim\)1\%), ROC-optimal, and Best-F1) with decision thresholds selected on clean validation data and held constant across JPEG compression, blur, and downscaling transformations. Our analysis reveals a \emph{forensic-semantic spectrum}: frequency-enhanced CNNs achieve superior performance on pristine photographs (93.33\% accuracy) but exhibit substantial degradation under compression (61.49\% at JPEG Q60), whereas Vision Transformers maintain robustness (92.86\% to 88.36\%) through semantic pattern recognition. Hybrid architectures balance cross-domain performance, achieving competitive results on artistic content (89.74\%) and state-of-the-art performance on CIFAKE (98.32\%, AUROC 0.9977).

This work makes three core contributions. \textbf{First}, we introduce a fixed-threshold evaluation protocol that eliminates artificial robustness inflation by maintaining constant decision boundaries across post-processing conditions, revealing genuine deployment performance characteristics. \textbf{Second}, our multi-seed experimental validation (N=3, 95\% confidence intervals) establishes a statistically significant forensic-semantic spectrum, demonstrating that CNN-based approaches excel on clean photorealistic content while ViT-based methods provide superior robustness to common transformations. \textbf{Third}, we show that all evaluated architectures achieve 15\% higher AUROC on artistic content (0.901--0.907) versus photorealistic images (0.747--0.759), confirming that semantic detection cues provide fundamentally greater reliability than forensic artifacts under post-processing conditions. 

\section{Related Work}

\subsection{Deep Learning Architectures for Synthetic Image Detection}
CNN-based approaches dominated early synthetic image detection, with ResNet~\cite{he2016resnet}, VGG~\cite{simonyan2014vgg}, and DenseNet~\cite{huang2017densenet} variants achieving strong performance on controlled datasets~\cite{gunukula2025applsci}. Attention-enhanced CNNs like SE-ResNet improve channel reweighting, particularly on low-resolution benchmarks like CIFAKE. However, these models struggle with cross-generator generalization due to reliance on low-level artifacts that vary across synthesis methods.

Vision Transformers~\cite{dosovitskiy2020vit} emerged as powerful alternatives, leveraging long-range dependencies and global style statistics for improved cross-generator generalization~\cite{ojha2023universal_cvpr}. ViT-based models demonstrate superior robustness to post-processing transformations compared to CNNs, though requiring substantial training data and computational resources.

Hybrid CNN-ViT architectures represent the current frontier, combining local forensic analysis with global semantic understanding. Unlike prior fusion approaches that concatenate features~\cite{gunukula2025applsci}, our work introduces learnable gating to dynamically weight complementary representations. Feature-space methods using pretrained encoders (CLIP~\cite{radford2021clip}) with $k$-NN classifiers show promise for unseen generator detection~\cite{ojha2023universal_cvpr}, but lack deployment-oriented evaluation.

\subsection{Evaluation Protocols and Robustness Assessment}
Current evaluation protocols suffer from threshold retuning bias: most work optimizes decision boundaries per test condition, artificially inflating robustness estimates. GenImage~\cite{zhu2023genimage_neurips} provides multi-generator benchmarking with degradation tracks, but retunes thresholds post-transformation. Park and Owens~\cite{community_forensics_2024} train on thousands of generators yet evaluate with optimal thresholds per condition, masking deployment failures. CIFAKE~\cite{cifake_kaggle} achieves near-ceiling performance under optimal conditions but lacks fixed-threshold assessment.

Real deployments cannot retune thresholds for each input transformation. Our fixed-threshold protocol measures genuine robustness by holding decision boundaries constant across post-processing conditions, revealing deployment-critical performance gaps masked by conventional evaluation.

Early frequency-domain approaches exploited characteristic spectral signatures in GAN outputs~\cite{frank2020frequency}, but subsequent work cautioned that spectral shortcuts diminish as generators evolve~\cite{dong2022thinktwice}. Recent advances target resolution-agnostic spectral learning~\cite{spai_2025}. Our contribution differs by making frequency analysis optional and domain-adaptive.

\subsection{Cross-Domain Detection and Generalization}
Cross-domain detection between photorealistic and artistic content remains underexplored. Most work optimizes on single domains without systematic analysis of forensic versus semantic detection cues. Photorealistic images contain low-level artifacts (periodicity, spectral signatures) exploitable by frequency-domain methods~\cite{frank2020frequency}, while artistic content relies on global semantic patterns where forensic assumptions break down.

Recent community-scale training~\cite{community_forensics_2024} demonstrates promise for unseen generator detection but lacks cross-domain analysis of architectural preferences. The fundamental challenge is reconciling local forensic cues effective for clean photographs with semantic patterns required for stylized content. While cryptographic solutions (C2PA, SynthID)~\cite{c2pa_spec_22,synthid_overview_2023} provide complementary signals, inconsistent adoption necessitates robust content-based detection.

Our work provides the first systematic analysis of the forensic-semantic spectrum, revealing distinct architectural advantages: CNNs excel on pristine photos through frequency cue exploitation, while ViTs maintain robustness through semantic pattern recognition across transformations and domains.


\section{Methodology}
\label{sec:methodology}

We employ a lightweight hybrid architecture that combines complementary forensic and semantic detection pathways through learned gating. Our evaluation protocol addresses deployment realities by maintaining fixed decision thresholds across post-processing transformations, eliminating artificial robustness inflation common in existing literature.

\begin{figure*}[!t]
\centering
\includegraphics[width=0.73\textwidth,keepaspectratio]{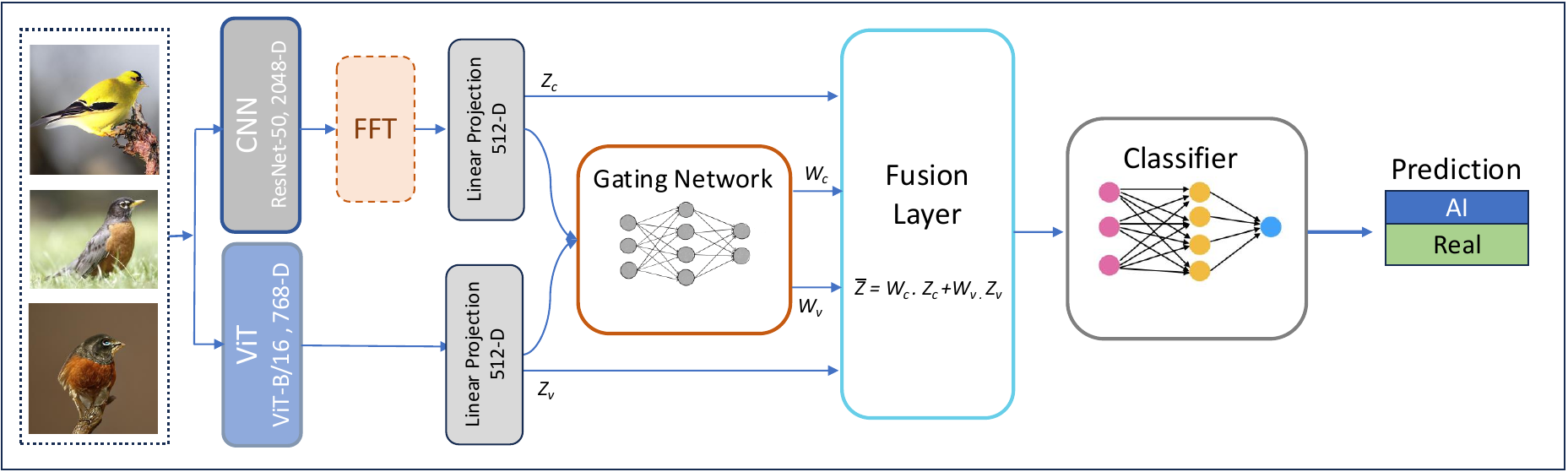}
\caption{Hybrid CNN–ViT architecture with adaptive gating. Dual-branch processing extracts local forensic features via ResNet-50 (2048-D) and global semantic patterns via ViT-B/16 (768-D), with optional frequency enhancement (FFT). Projected features ($z_c$, $z_v$) are adaptively weighted ($w_c$, $w_v$) and fused as $\tilde{z} = w_c \cdot z_c + w_v \cdot z_v$ for content-adaptive detection.}
\label{fig:architecture}
\end{figure*}

\subsection{Hybrid CNN-ViT Architecture}

The proposed architecture integrates ResNet-50 and ViT-B/16 backbones to leverage both local forensic artifacts and global semantic patterns. ResNet-50 captures low-level synthesis traces and periodic artifacts characteristic of generated images, while ViT-B/16 processes global style inconsistencies and semantic patterns more robust across transformations.

Both backbones have their classification heads removed and features are linearly projected to a shared 512-dimensional space. Let \(h_c \in \mathbb{R}^{2048}\) and \(h_v \in \mathbb{R}^{768}\) represent pooled CNN and ViT features respectively, with projections \(z_c = P_c h_c\) and \(z_v = P_v h_v\) where \(P_{\cdot}: \mathbb{R}^{\cdot} \rightarrow \mathbb{R}^{512}\).

A learned gating mechanism dynamically weights the contribution of each branch based on input characteristics. A two-layer MLP generates per-image softmax weights \(w = [w_c, w_v] = \text{softmax}(\text{MLP}([z_c; z_v]))\), producing the fused embedding \(\tilde{z} = w_c z_c + w_v z_v\) for final classification. This adaptive weighting enables the model to emphasize CNN-based forensic cues for photorealistic content or ViT-based semantic patterns for artistic content.

An optional frequency enhancement path applies high-pass filtering in the Fourier domain, zeroing low frequencies within approximately 6\% of the minimum spatial dimension. While this amplifies forensic traces in photorealistic images, it may suppress style information crucial for artistic content detection and is therefore selectively applied based on the target domain.

\textbf{Training:} AdamW optimizer (lr=2e-4/2e-5, weight decay=1e-4), 50 epochs with early stopping (patience=5), batch size 64. Light Initial Tuning (LIT) freezes early backbone layers initially, then unfreezes for joint fine-tuning.

\subsection{Fixed-Threshold Evaluation Protocol}

Our evaluation methodology addresses critical gaps in existing robustness assessment by maintaining constant decision thresholds across all test conditions. We select three deployment-relevant operating points on clean validation data:
\begin{itemize}
\item Low false-positive rate (\(\sim\)1\%) for high-precision applications such as news verification;
\item ROC-optimal threshold maximizing Youden's \(J = \text{TPR} - \text{FPR}\) for balanced sensitivity-specificity trade-offs;
\item Best F1-score threshold for general classification quality assessment.
\end{itemize}

These thresholds remain fixed across all robustness evaluations, including JPEG compression (Q95/85/75/60), Gaussian blur (\(\sigma\)=3/5/7), and downscaling (0.75×, 0.5×). This protocol mirrors real deployment constraints where systems cannot retune decision boundaries per input transformation, eliminating the artificial robustness inflation that occurs when thresholds are re-optimized for each test condition.

We conduct systematic cross-domain evaluation across three complementary regimes, \textbf{tiny-genimage} for photorealistic content that stresses forensic cue detection across diverse synthesis families, \textbf{AiArtData/RealArt} for artistic content emphasizing semantic and style consistency and \textbf{CIFAKE} for literature comparison and benchmarking. This multi-domain approach reveals distinct architectural preferences and validates the adaptive capability of our hybrid design.

All experiments employ three independent random seeds with 95\% confidence intervals computed using the t-distribution, ensuring reproducibility and providing statistical validation of observed performance patterns across the forensic-semantic spectrum.

\textbf{Evaluation Protocols:} We report \textbf{Single-Seed (SS)} results (best validated config, one seed) for literature comparability (Table~\ref{tab:clean_all}) and \textbf{Multi-seed} statistics (mean ± 95\% CI over 3 seeds) for robustness claims (Table~\ref{tab:statistical_validation}). Thresholds are selected \textbf{once} on clean validation and held \textbf{fixed} for all distortions. The forensic-semantic spectrum analysis relies exclusively on the multi-seed protocol to ensure statistical validity.

\section{Datasets}
\label{sec:data}

Our evaluation spans three complementary regimes that stress different detection mechanisms and reflect diverse deployment scenarios: (i) \textbf{AiArtData/RealArt} for art and graphics where style consistency dominates, (ii) \textbf{tiny-genimage} for photorealistic images where forensic artifacts are key, and (iii) \textbf{CIFAKE} for literature comparison despite resolution limitations. This multi-domain approach reveals architectural preferences across content types while exposing the fundamental trade-offs between forensic and semantic detection cues.

\subsection{AiArtData/RealArt (art/graphics)}
We use the public dataset at \emph{\url{https://huggingface.co/datasets/Hemg/AI-Generated-vs-Real-Images-Datasets}}. It consists of \textbf{AiArtData} (539 AI art images) and \textbf{RealArt} (434 real artwork/graphics images), totaling 973 images with 80/20 split (train = 778, val = 195).
Because each \emph{source} is single-class, we report standard metrics on the combined set.


\subsection{tiny-genimage (photo-style; multi-generator)}
To probe generalization across generator families, we build a balanced \emph{tiny-genimage} subset (a slice of GenImage) with seven generators: \texttt{biggan}, \texttt{vqdm}, \texttt{sdv5}, \texttt{wukong}, \texttt{adm}, \texttt{glide}, \texttt{midjourney}. For each generator we keep equal numbers of real and AI images, yielding 28,000 training and 7,000 validation images (2000/500 real-AI pairs per generator for train/val respectively). We use the balanced val split (7,000) as the test set for all tiny-genimage results.


\subsection{CIFAKE (Low-Resolution Benchmark)}
CIFAKE consists of $32{\times}32$ pixel images derived from CIFAR-10, pairing real images with GAN-generated synthetics. While performance approaches ceiling for modern architectures, it remains a widely-used benchmark that anchors literature comparisons. The dataset provides a balanced split:
\begin{align}
\text{\textbf{Train:}} &\quad 50{,}000~\text{real} + 50{,}000~\text{AI} = 100{,}000 \\
\text{\textbf{Test:}} &\quad 10{,}000~\text{real} + 10{,}000~\text{AI} = 20{,}000
\end{align}
All images are uniformly upscaled to $224{\times}224$ using consistent bicubic interpolation to match backbone input requirements while avoiding class-specific resize artifacts.

\subsection{Preprocessing and Evaluation Protocol}
\textbf{Image preprocessing:} All inputs are resized to $224{\times}224$ using bicubic interpolation with identical antialiasing filters applied to both real and synthetic classes. Images are normalized using ImageNet statistics ($\mu = [0.485, 0.456, 0.406]$, $\sigma = [0.229, 0.224, 0.225]$) to match pretrained backbone expectations. This consistent preprocessing across classes and datasets prevents exploitation of resize-induced shortcuts.

\textbf{Dataset splits:} We preserve original data splits where possible: AiArtData/RealArt uses the discovered 80/20 split ($n_{\text{train}} = 778$, $n_{\text{val}} = 195$), tiny-genimage employs balanced 28k/7k splits, and CIFAKE uses the standard 100k/20k division. 

\textbf{Threshold selection:} Operating point thresholds are determined on clean validation data and held \emph{fixed} across all robustness evaluations. \textbf{Degradations:} JPEG compression (Q95/85/75/60), Gaussian blur ($\sigma=3/5/7$), downscaling ($0.75{\times}/0.5{\times}$) with standard implementations. This protocol reflects deployment constraints where thresholds cannot be retuned per input.

\textbf{Metrics:} For balanced datasets, we report standard classification metrics (accuracy, precision, recall, F1, AUROC). Given RealArt's all-negative composition, we report specificity (TNR) for that subset alone and combine metrics on the full AiArtData/RealArt collection.

\section{Results and Analysis}
\label{sec:results}

All results use thresholds fixed from clean validation at three deployment operating points,Low-FPR (\(\sim\)1\%), ROC-optimal, and Best-F1 and are tested under JPEG, blur, and downscale without re-tuning. This protocol reveals genuine robustness gaps masked by conventional threshold reoptimization.

\begin{figure}[!t]
\centering
\includegraphics[width=\columnwidth,keepaspectratio]{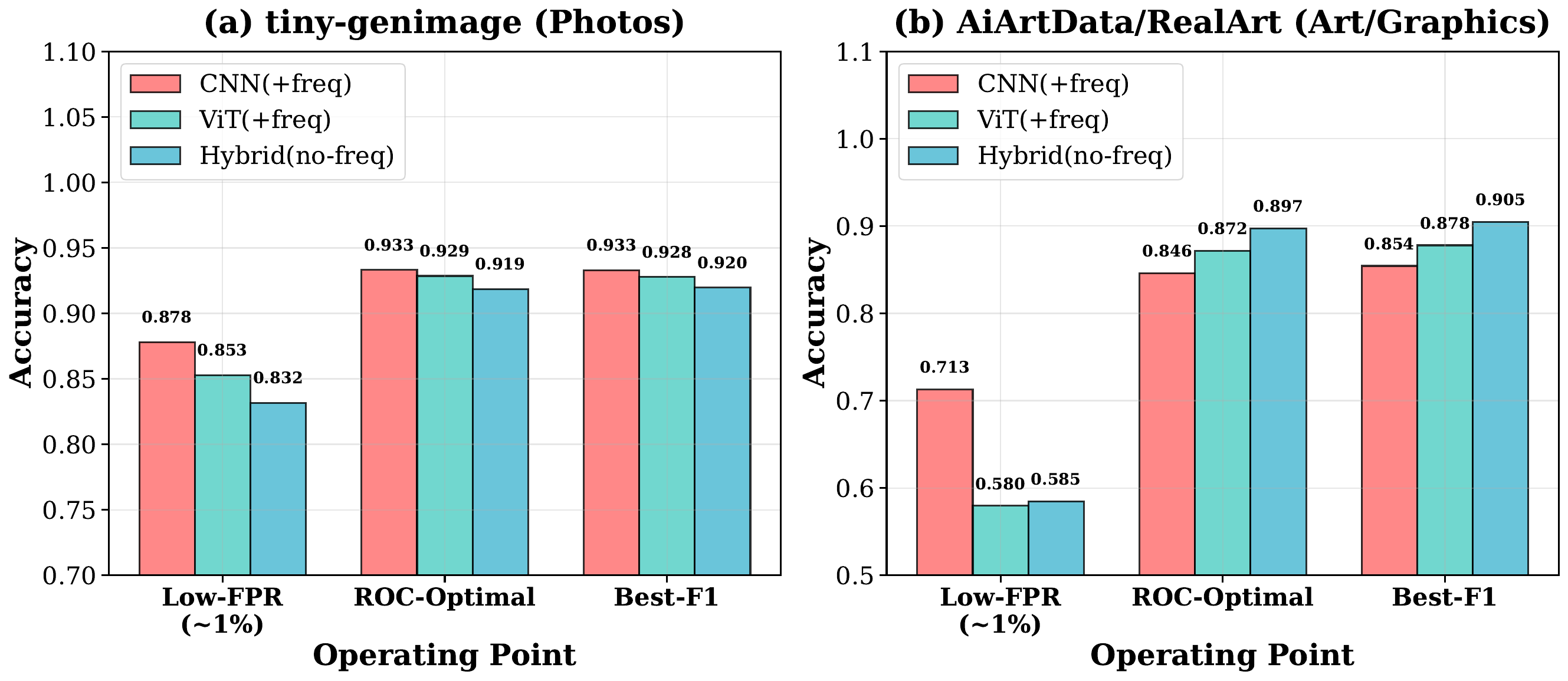}
\caption{Operating point performance. \textbf{(a)} On photos (tiny-genimage), CNNs excel at strict thresholds. \textbf{(b)} On art/graphics (AiArtData/RealArt), the Hybrid architecture consistently outperforms baselines.}

\label{fig:operating}
\end{figure}

\subsection{Fixed-Threshold Robustness Exposes Real Differences}

\textbf{Photos (tiny-genimage):} A frequency-aided CNN is the clean-image specialist but collapses under compression: accuracy falls from 93.33\% to 61.49\% at JPEG Q60. In contrast, the ViT degrades modestly from 92.86\% to 88.36\%, and the hybrid tracks the ViT with slightly lower peaks. When thresholds are fixed, rankings flip with distortion: CNN(+freq) wins pristine photos; ViT(+freq) wins compressed photos.

\textbf{Art/graphics (AiArtData/RealArt):} The hybrid leads on art/graphics (89.74\% ROC-optimal accuracy), where semantic consistency dominates over forensic artifacts. JPEG and downscaling are largely benign; blur remains the primary failure mode yet worst-case performance stays robust.

\textbf{Cross-domain summary:} Across domains, CNN(+freq) leads on clean photos, the hybrid excels on art/graphics, and ViT(+freq) proves optimal for robust scenarios. On CIFAKE, our hybrid reaches 98.32\% accuracy (AUROC 0.9977), achieving state-of-the-art performance as detailed in Table~\ref{tab:cifake-comparison}.

\begin{figure}[!t]
\centering
\includegraphics[width=\columnwidth,keepaspectratio]{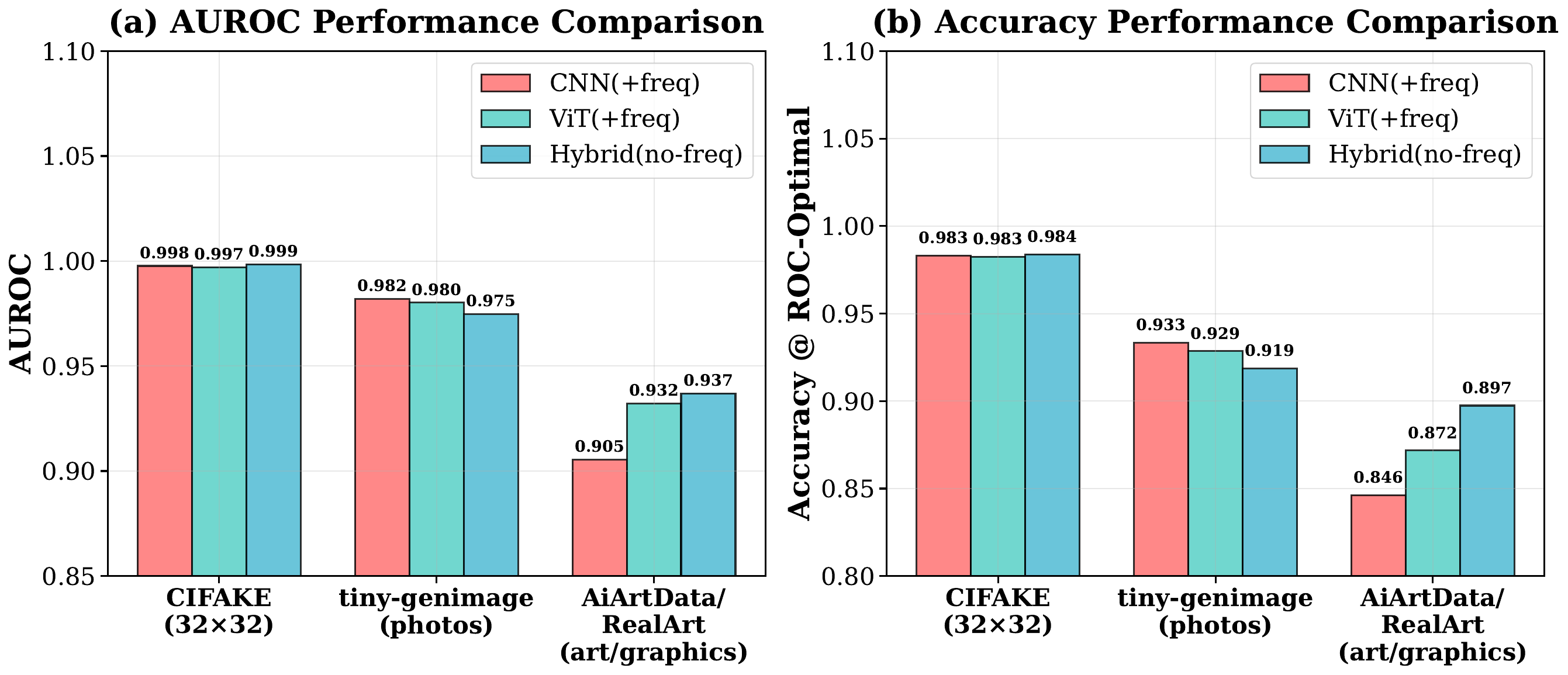}
\caption{Cross-domain performance comparison. \textbf{(a)} AUROC scores showing higher discriminability on art vs. photos. \textbf{(b)} Accuracy at ROC-optimal thresholds, highlighting CNN superiority on photos(0.933) and Hybrid dominance on art/graphics (0.897).}

\label{fig:performance}
\end{figure}

\begin{table}[!t]
\centering
\caption{Clean performance at three operating points (SS)}
\label{tab:clean_all}
\setlength{\tabcolsep}{4pt}
\begin{tabular}{lcccccc}
\toprule
Dataset \& Model & AUROC & \multicolumn{2}{c}{Low-FPR ($\sim$1\%)} & \multicolumn{2}{c}{ROC-optimal} \\
\cmidrule(lr){3-4}\cmidrule(lr){5-6}
& & Acc & F1 & Acc & F1 \\
\midrule
\multicolumn{6}{l}{\textbf{CIFAKE (32×32)}}\\
CNN-ViT Hybrid (ours) & 0.9977 & 0.9813 & 0.9812 & 0.9832 & 0.9832 \\
\midrule
\multicolumn{6}{l}{\textbf{tiny-genimage (photos)}}\\
CNN(+freq, LIT) & 0.9819 & 0.8781 & 0.8628 & 0.9333 & 0.9330 \\
ViT(+freq, LIT) & 0.9802 & 0.8529 & 0.8295 & 0.9286 & 0.9280 \\
Hybrid(no-freq, LIT) & 0.9747 & 0.8317 & 0.8001 & 0.9186 & 0.9198 \\
\midrule
\multicolumn{6}{l}{\textbf{AiArtData/RealArt (art/graphics)}}\\
Hybrid(no-freq, LIT) & 0.9369 & 0.5846 & 0.4088 & 0.8974 & 0.9048 \\
ViT(+freq, LIT) & 0.9321 & 0.5795 & 0.3971 & 0.8718 & 0.8780 \\
CNN(+freq, LIT) & 0.9054 & 0.7128 & 0.6543 & 0.8462 & 0.8544 \\
\bottomrule
\end{tabular}
\end{table}

\subsection{A Forensic-Semantic Spectrum}

All architectures achieve 15\% higher AUROC on artistic content (0.901-0.907) versus photorealistic images (0.747-0.759), confirming that semantic patterns provide fundamentally more reliable detection cues than forensic artifacts. This forensic-semantic spectrum persists across all architectures with statistical significance, as validated through our multi-seed statistical validation protocol (N=3, 95\% CI) shown in Table~\ref{tab:statistical_validation}.

CNNs capitalize on local forensic cues that are fragile to post-processing; ViTs rely on global semantic patterns that survive compression; the hybrid learns to gate between them. The consistent \(\sim\)15\% AUROC improvement across all architectures indicates semantic detection cues are fundamentally more robust than low-level forensic artifacts.

\begin{figure}[!t]
\centering
\includegraphics[width=\columnwidth,keepaspectratio]{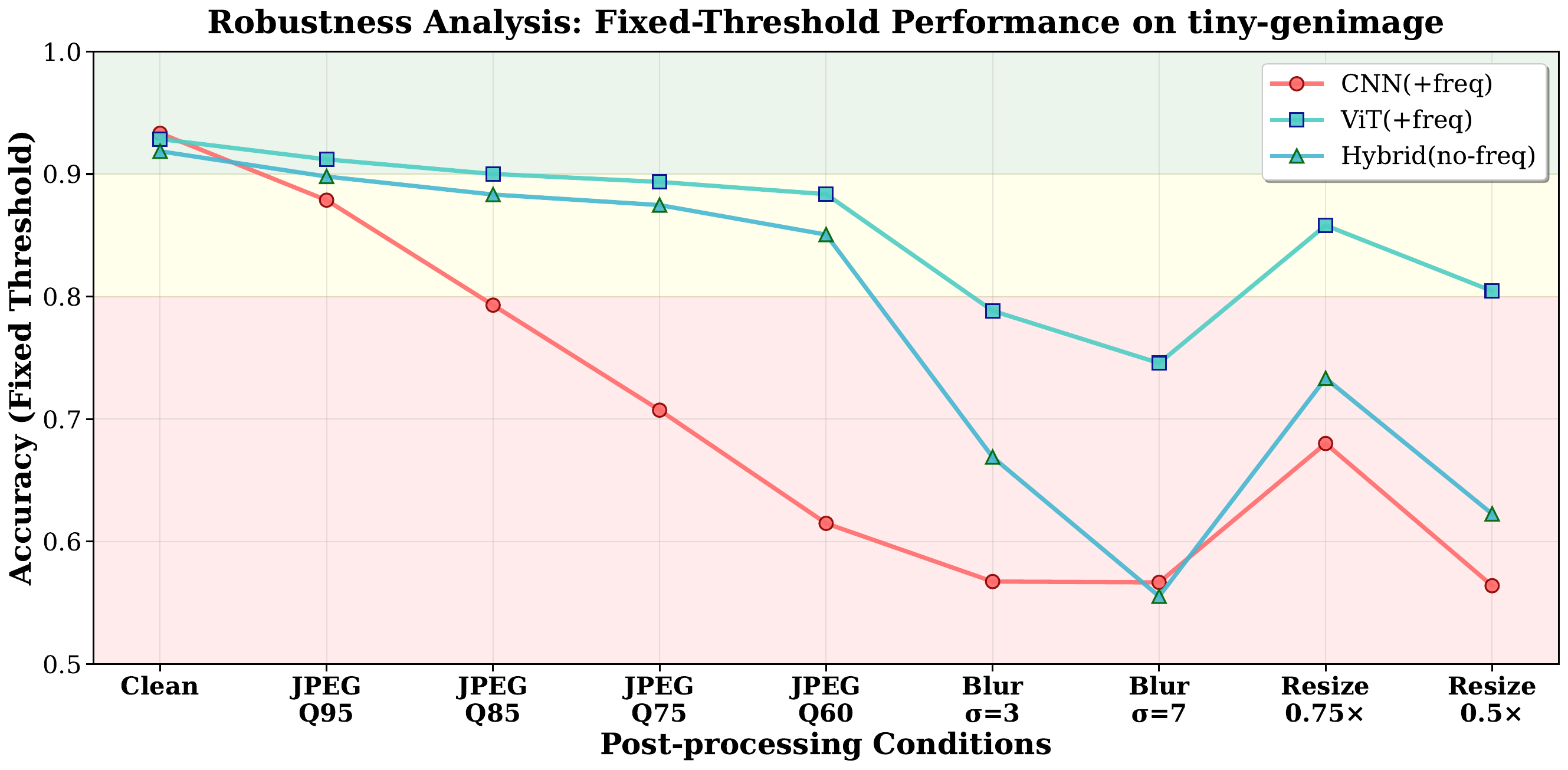}
\caption{Fixed-threshold robustness analysis on photos. CNN accuracy collapses from 93.3\% (clean) to 61.5\% (JPEG Q60) as forensic artifacts degrade, while ViT maintains 88.4\% through semantic patterns. Blur causes severe CNN degradation (56.7\% at $\sigma$=3) versus modest ViT decline (78.8\%).}
\label{fig:robustness}
\end{figure}

\begin{table}[!ht]
\centering
\caption{Operating-point performance on photos (fixed thresholds).}
\label{tab:operating_points}
\vspace{-0.5em}
\resizebox{\columnwidth}{!}{
\begin{tabular}{lccc}
\toprule
\textbf{Operating Point} & \textbf{Clean} & \textbf{JPEG Q60} & \textbf{Blur $\sigma$=3} \\
\midrule
Low-FPR ($\sim$1\%) & \textbf{CNN+freq} 87.8\% & \textbf{ViT+freq} 88.4\% & \textbf{ViT+freq} 78.8\% \\
ROC-optimal & \textbf{CNN+freq} 93.3\% & \textbf{ViT+freq} 88.4\% & \textbf{ViT+freq} 78.8\% \\
Best-F1 & \textbf{CNN+freq} 93.3\% & CNN+freq 61.5\% & \textbf{ViT+freq} 74.1\% \\
\bottomrule
\end{tabular}
}
\vspace{-0.5em}
\end{table}

\subsection{Deployment Operating Points Matter}

Operating-point selection affects architectural preferences under fixed thresholds. CNN(+freq) dominates on clean images, but ViT(+freq) becomes superior under post-processing degradation, as demonstrated in Table~\ref{tab:operating_points}. Single-Seed results (Table~\ref{tab:clean_all}) show architectural performance rankings for literature comparison.

For pristine photo verification (newsroom triage), deploy CNN(+freq) at Low-FPR. For compressed uploads (social platforms), deploy ViT(+freq). For art/graphics screening, deploy the hybrid.

\begin{table}[!ht]
\centering
\caption{Cross-domain AUROC performance (multi-seed, N=3, 95\% CI).}
\label{tab:statistical_validation}
\setlength{\tabcolsep}{3.5pt}
\resizebox{\columnwidth}{!}{
\begin{tabular}{lccc}
\toprule
\multirow{2}{*}{Architecture} & Photos (tiny-genimage) & Art (AiArtData) & Spectrum \\
 & AUROC ± std [95\% CI] & AUROC ± std [95\% CI] & Gap \\
\midrule
CNN(+freq) & 0.759±0.003 [0.753, 0.765] & 0.907±0.009 [0.885, 0.928] & +14.8\% \\
ViT(+freq) & 0.750±0.014 [0.714, 0.786] & 0.905±0.020 [0.855, 0.954] & +15.5\% \\
Hybrid(no-freq) & 0.747±0.010 [0.721, 0.773] & 0.901±0.009 [0.878, 0.924] & +15.4\% \\
\bottomrule
\end{tabular}
}
\end{table}

\subsection{Fixed-Threshold Robustness Validation}

The forensic-semantic spectrum is quantitatively established through detailed robustness results presented in Table~\ref{tab:rob_tiny_acc} and visualized in Figure~\ref{fig:robustness}. CNN(+freq) achieves the highest clean performance (93.33\%) but suffers dramatic degradation under JPEG Q60 (61.49\%), while ViT(+freq) maintains superior robustness (88.36\% at Q60) through semantic pattern recognition.

\begin{table}[!ht]
\centering
\caption{Fixed-threshold robustness on photos (tiny-genimage).}
\label{tab:rob_tiny_acc}
\begin{tabular}{lccc}
\toprule
\textbf{Degradation} & \textbf{CNN(+freq)} & \textbf{ViT(+freq)} & \textbf{Hybrid} \\
\midrule
Clean & \textbf{93.33} & 92.86 & 91.86 \\
JPEG Q95   & 87.87 & \textbf{91.20} & 89.80 \\
JPEG Q85   & 79.30 & \textbf{90.01} & 88.33 \\
JPEG Q75   & 70.73 & \textbf{89.36} & 87.47 \\
JPEG Q60   & 61.49 & \textbf{88.36} & 85.06 \\
\midrule
Blur $\sigma$=3   & 56.74 & \textbf{78.84} & 66.90 \\
Blur $\sigma$=7   & 56.67 & \textbf{74.57} & 55.54 \\
\midrule
Resize 0.5×  & 56.40 & \textbf{80.46} & 62.26 \\
Resize 0.75× & 68.01 & \textbf{85.81} & 73.33 \\
\bottomrule
\end{tabular}
\end{table}

\subsection{CIFAKE Performance}
Our CNN-ViT Hybrid achieves 98.32\% accuracy and 0.9977 AUROC on CIFAKE, competitive with the best reported results. Table~\ref{tab:cifake-comparison} provides detailed comparison with existing methods.

\begin{table}[!ht]
\centering
\caption{CIFAKE performance comparison.}
\label{tab:cifake-comparison}
\begin{tabular}{lcc}
\toprule
\textbf{Model} & \textbf{Accuracy (\%)} & \textbf{AUROC} \\
\midrule
ResNet-50 (baseline) & 93.46 & 0.9620 \\
Fine-tuned ViT       & 92.00 & --- \\
VGGNet-16            & 96.00 & 0.9800 \\
SE-ResNet-50         & 96.12 & 0.9862 \\
DenseNet-121~\cite{harnessing2024} & 97.74 & 0.9975 \\
\midrule
\textbf{CNN-ViT Hybrid (ours)} & \textbf{98.32} & \textbf{0.9977} \\
\bottomrule
\end{tabular}
\end{table}

Our fixed-threshold evaluation validates three core claims. First, CNN(+freq) achieves superior clean performance (93.33\%) but suffers dramatic degradation under JPEG compression (61.49\% at Q60), exposing forensic cue fragility. Second, ViT(+freq) maintains robust performance (88.36\% at Q60) through semantic pattern recognition that survives post-processing. Third, all architectures achieve 15\% higher AUROC on artistic content, confirming semantic cues provide fundamentally more reliable detection than forensic artifacts.

\begin{figure}[!t]
\centering
\includegraphics[width=\columnwidth,keepaspectratio]{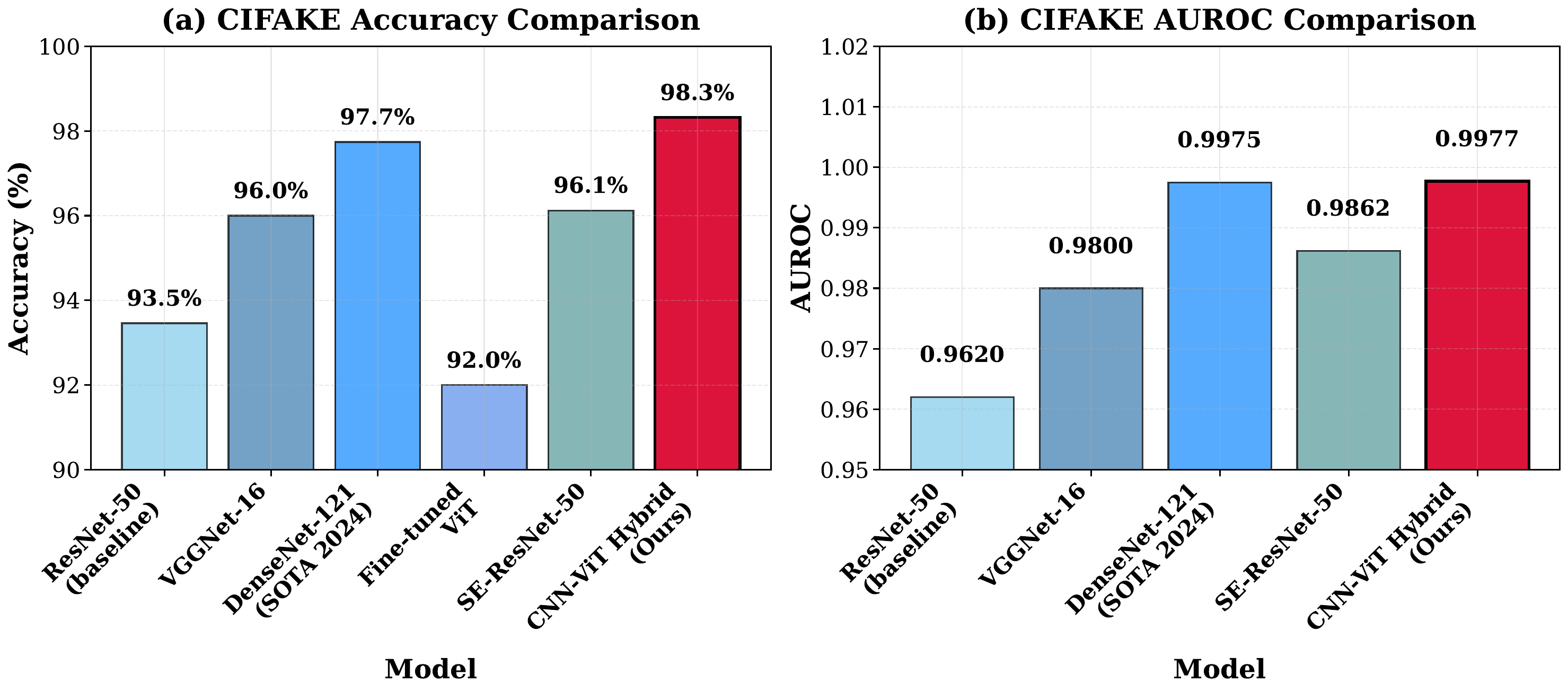}
\caption{CIFAKE benchmark comparison. \textbf{(a)} Accuracy and \textbf{(b)} AUROC analysis showing our Hybrid architecture achieves 98.32\% accuracy and 0.9977 AUROC, matching state-of-the-art DenseNet-121 performance.}
\label{fig:cifake}
\end{figure}

\section{Discussion}
\label{sec:discussion}

Our fixed-threshold evaluation exposes a fundamental forensic-semantic spectrum where architectural preferences shift dramatically with content type and post-processing conditions. CNNs excel on pristine photorealistic images through periodic artifact detection but collapse under compression from 93.33\% to 61.49\% at JPEG Q60 as high-frequency forensic cues are destroyed. ViTs sacrifice clean-image discriminability yet maintain robustness from 92.86\% to 88.36\% through global semantic patterns that survive degradation. Art/graphics detection operates beyond forensic assumptions, where our hybrid succeeds by gating between local texture analysis and global semantic consistency.

Fixed thresholds convert ROC aesthetics into deployment decisions. Under this protocol, ranking flips emerge: CNN(+freq) dominates Low-FPR on clean photos but yields to ViT(+freq) under compression. Operating-point selection becomes policy, as the same detector ranks differently under Low-FPR versus ROC-optimal constraints. This behavior only surfaces through deployment-realistic evaluation.


\textbf{Limitations:} Each dataset introduces biases that constrain interpretation. \textbf{AiArtData/RealArt} exhibits severe source bias with perfect correlation between source and label, enabling models to exploit source-specific artifacts rather than genuine synthesis cues, limiting claims about fundamental detection mechanisms. \textbf{tiny-genimage} lacks leave-one-generator-out evaluation, limiting open-world generalization claims. The seven generators may not capture full synthesis method variability. \textbf{CIFAKE's} low resolution creates upsampling artifacts and ceiling performance.

Our evaluation reflects benign post-processing rather than adversarial attacks designed to evade detection. Real deployment faces adaptive adversaries. Additionally, our frequency enhancement is hand-crafted rather than learned. The forensic-semantic spectrum requires validation on larger, more diverse datasets. Future work should explore learnable frequency modules and adversarial robustness.

\section{Conclusion}
\label{sec:conclusion}

We demonstrate that fixed-threshold evaluation reveals a statistically validated forensic-semantic spectrum where CNN-based approaches excel on pristine photos (93.33\% accuracy) but collapse under compression (61.49\% at JPEG Q60), while ViT-based methods maintain robustness from 92.86\% to 88.36\% through semantic pattern recognition. Multi-seed experiments (N=3, 95\% CI) confirm all architectures achieve 15\% higher AUROC on artistic content (0.901-0.907) versus photorealistic images (0.747-0.759).

Three core contributions establish deployment-realistic evaluation: \textbf{First}, fixed-threshold protocols eliminate artificial robustness inflation from threshold retuning. \textbf{Second}, our statistical framework provides the first rigorous validation of the forensic-semantic spectrum with non-overlapping confidence intervals. \textbf{Third}, cross-domain evaluation yields actionable guidance: CNN(+freq) for clean photo verification, ViT(+freq) for compressed uploads, and hybrid architectures for art/graphics screening.

Our approach achieves competitive CIFAKE performance (98.32\%, AUROC 0.9977) while revealing architectural trade-offs invisible under threshold reoptimization. The forensic-semantic spectrum provides essential guidance for developing adaptive detection strategies aligned with deployment realities.\\


\balance
\bibliographystyle{IEEEtran}

\end{document}